\documentclass[11pt,a4paper]{article}

\usepackage[utf8]{inputenc}
\usepackage[T1]{fontenc}
\usepackage{times}
\usepackage{graphicx}
\usepackage{booktabs}
\usepackage{amsmath}
\usepackage{amssymb}
\usepackage{hyperref}
\usepackage[margin=1in]{geometry}
\usepackage{caption}
\usepackage{subcaption}
\usepackage{enumitem}
\usepackage{float}
\usepackage{color}
\usepackage{listings}
\usepackage{xcolor}
\usepackage{multirow}
\usepackage{tikz}
\usetikzlibrary{arrows.meta, positioning, shapes.geometric}
\usepackage{array}
\usepackage[numbers]{natbib}
\usepackage{authblk}
\newenvironment{keywords}{\par\medskip\noindent\textbf{Keywords:} \itshape}{\par\medskip}

\definecolor{codebg}{rgb}{0.95,0.95,0.95}
\title{\textbf{The Public Discourse Corpus (PDC): \\
A Speaker-Attributed Dataset for Valence and \\
Epistemic Modality with Target Speaker Participation}}

\author[1]{Bo Chen}
\affil[1]{Institute of Computing Technology, Chinese Academy of Sciences}

\date{}

\begin{document}
\maketitle

\begin{abstract}
We introduce the \textbf{Public Discourse Corpus (PDC)}, the first dataset of public-figure interview speech jointly annotated for affective valence and epistemic modality. The corpus contains 998 videos from 100 speakers across seven professional domains, yielding 186,642 sentences (3.1 million words) after sentence segmentation and filtering. To ensure that all retained videos contain analyzable speech from the intended speaker, we introduce \textbf{Target Speaker Participation (TSP)}---a five-category annotation taxonomy with documented inter-annotator reliability ($\kappa = 0.616$)---as a key methodological contribution that any corpus construction project can adopt. Target-speaker turns are separated from interviewer and third-party speech through an \textbf{audio-first diarization pipeline} combining local Whisper ASR with pyannote speaker separation, released as an open-source implementation. We release the annotated corpus, the annotation tools, the cross-provider validation sample, and the complete processing pipeline. The dataset is available at \url{https://huggingface.co/datasets/ictchenbo/public-discourse-corpus}.
\end{abstract}


\section{Introduction}

Computational social science (CSS) increasingly relies on NLP to measure psychological and rhetorical constructs---sentiment, certainty, emotion, stance---at scale~\cite{bail_2024, ziems_2024}. The dominant paradigms are well-established: lexicon-based methods (LIWC~\cite{liwc}, VADER~\cite{vader}), transformer-based classifiers (BERT~\cite{bert}, RoBERTa~\cite{roberta}), and large language models (LLMs)~\cite{gilardi_2023} for zero-shot annotation.

However, a critical gap exists in available annotated resources. Existing corpora for public discourse provide either sentiment labels or modality labels, but no dataset combines both dimensions on the same speaker-attributed utterances (see Table~\ref{tab:existing_corpora}). This is a consequential omission: the relationship between \textit{what} a speaker says (valence) and \textit{how} assertively or cautiously they say it (modality) cannot be studied unless both dimensions are annotated on the same units of speech with known speaker identity. For example, researchers cannot currently ask whether central bankers hedge more when expressing negative versus positive forecasts, or whether politicians across different parties differ in how emphatically they frame negative claims about opponents. These questions require simultaneous valence and modality labels linked to speaker metadata---a resource that does not exist.

Building such a corpus from YouTube presents a practical challenge: automatic video retrieval returns substantial amounts of content \textit{about} a public figure without containing speech \textit{by} them. In our manual review of 122 randomly sampled candidate videos, 47\% were third-party content (news reports, panel discussions, documentary narration) rather than interviews featuring the target speaker. Without systematic verification, nearly half of automatically retrieved content would be misattributed, contaminating any downstream analysis.

We address this gap by introducing the \textbf{Public Discourse Corpus (PDC)}, a speaker-attributed dataset of 998 videos from 100 speakers across seven professional domains, with 186,642 sentences jointly annotated for affective valence and epistemic modality. The PDC is the primary contribution. To ensure that all retained videos contain analyzable speech from the intended speaker, we develop \textbf{Target Speaker Participation (TSP)}, a systematic five-category taxonomy for verifying target-speaker presence in automatically retrieved videos. TSP is a key methodological contribution in its own right: it addresses a general problem (``content about vs.\ speech by'') that affects any corpus built from media sources, and is validated through documented inter-annotator agreement ($\kappa = 0.616$). Target-speaker turns are separated from interviewer and third-party speech through an \textbf{audio-first diarization pipeline} that uses acoustic speaker-turn evidence rather than text-only heuristics; the pipeline is released as open-source software to support replication and reuse.

Our contributions are:

\begin{enumerate}[nosep]
  \item \textbf{The Public Discourse Corpus (PDC)} --- \textit{primary contribution:} the first public discourse corpus jointly annotated for affective valence and epistemic modality, spanning 998 videos from 100 speakers across seven domains (186,642 sentences). The corpus enables research on the interaction between evaluative language and epistemic commitment that was previously impossible at this scale and domain diversity.
  \item \textbf{Target Speaker Participation (TSP)} --- \textit{key methodological contribution:} a reusable taxonomy and annotation protocol for verifying target-speaker presence during corpus construction from video sources, with documented inter-annotator reliability ($\kappa = 0.616$ for five-way classification) and released annotation guidelines.
  \item \textbf{Audio-first construction pipeline} --- \textit{implementation contribution:} an open-source, fully local pipeline combining Whisper large-v3, pyannote/speaker-diarization-3.1, and LLM-based dual-dimension annotation, enabling replication of the PDC construction process.
\end{enumerate}

\paragraph{How researchers can use the PDC.}
The corpus is released as three joinable JSONL files (see Section~\ref{sec:dataset_structure}). Researchers can filter by domain, TSP category, or speaker to construct analysis subsets; compute per-speaker valence and modality profiles; and analyze cross-dimension relationships at multiple levels of aggregation. A stratified cross-provider validation sample (Section~\ref{sec:llm_agreement}) enables users to bound label uncertainty in their own analyses. Section~\ref{sec:illustrative} provides a concrete example.

\section{Related Work}

\subsection{Public Discourse Datasets}

Table~\ref{tab:existing_corpora} situates the PDC against existing resources. CMU-MOSEI~\cite{mosei} provides sentiment annotations on YouTube monologues, IEMOCAP~\cite{iemocap} and MELD~\cite{meld} annotate acted or scripted emotional speech, MPQA~\cite{mpqa} covers sentiment and subjectivity on news text, and FactBank~\cite{factbank} provides factuality labels. No existing corpus provides dual-dimension (valence + epistemic modality) annotation for natural public discourse at any meaningful scale, and none address the target-speaker attribution problem that the PDC's TSP taxonomy solves.

\begin{table}[H]
\centering
\caption{Comparison of existing corpora with the PDC.}
\label{tab:existing_corpora}
\resizebox{\textwidth}{!}{%
\begin{tabular}{lccccc}
\toprule
\textbf{Corpus} & \textbf{Size} & \textbf{Modality} & \textbf{Annotations} & \textbf{Discourse} & \textbf{V+M?} \\
\midrule
CMU-MOSEI~\cite{mosei} & 23.5K sent. & Video+Audio+Text & Sentiment (1--5) & YouTube monologue & $\times$ \\
IEMOCAP~\cite{iemocap} & 10K utt. & Audio+Video & Emotion cats. & Acted dyadic & $\times$ \\
MELD~\cite{meld} & 13.7K utt. & Video+Audio+Text & Emotion+Sent. & Scripted TV & $\times$ \\
MPQA~\cite{mpqa} & 15K sent. & Text & Sent.+Subjectivity & News & $\times$ \\
FactBank~\cite{factbank} & 10K events & Text & Factuality/Cert. & News & $\times$ (M only) \\
\midrule
\textbf{PDC (this work)} & \textbf{998 videos} & \textbf{Audio+Text} & \textbf{Valence + Modality} & \textbf{Interviews (100 spk.)} & \textbf{$\checkmark$} \\
\bottomrule
\end{tabular}%
}
\end{table}

The PDC is distinguished by its scale, dual-dimension annotation, and 100-speaker cross-domain coverage. Beyond the dataset itself, the construction methodology---specifically the TSP annotation protocol---offers a replicable approach for building target-speaker corpora from video sources.

\subsection{Speaker Diarization for Corpus Construction}

Speaker diarization---the task of partitioning audio into speaker-homogeneous segments---has seen substantial progress. Neural approaches (pyannote~\cite{pyannote}, NeMo,\footnote{NVIDIA NeMo: \url{https://github.com/NVIDIA/NeMo}} WhisperX\footnote{WhisperX: \url{https://github.com/m-bain/whisperX}}) and cloud services (AssemblyAI\footnote{AssemblyAI: \url{https://www.assemblyai.com}}) provide high-accuracy diarization for clean audio. However, this literature targets \textit{diarization accuracy} as an end in itself, typically benchmarking on datasets like AMI or VoxConverse. Our contribution is different: we evaluate diarization not as a standalone task but as a \textit{corpus-construction component}. The key question is whether audio evidence (pause duration, pitch shifts, voice characteristics) provides more reliable speaker attribution than text-only inference for the specific purpose of building an annotated dataset. Recent work by Chen~\cite{chen_instability_2026} quantified the downstream impact of pipeline choices on valence--modality measurements, motivating our choice of an audio-first route.

\subsection{LLM-Based Annotation}

The emergence of LLMs has enabled zero-shot annotation at scale. Gilardi, Alizadeh, and Kubli~\cite{gilardi_2023} demonstrated that ChatGPT outperforms crowd workers on multiple text-annotation tasks. Ziems et al.~\cite{ziems_2024} surveyed the broader landscape of LLMs in CSS, identifying prompt sensitivity and provider-specific biases as key challenges. In the domain of stance and modality, Kim, Muhn, and Nikolaev~\cite{kim_llm_financial_2024} showed that LLMs can extract modality-adjacent constructs from corporate text at scale. The PDC adopts LLM-based annotation for both valence and modality, with cross-provider validation to document provider-specific variation (Section~\ref{sec:llm_agreement}). Our contribution is not a new annotation model but a \textit{paired annotation scheme}---valence and modality assigned simultaneously to the same utterances---applied to a speaker-attributed corpus built through systematic preprocessing.

\section{Corpus Construction Pipeline}

\begin{figure}[H]
\centering
\begin{tikzpicture}[
    node distance=8mm and 0mm,
    process/.style={draw, rounded corners, minimum width=36mm, minimum height=7mm, align=center, font=\small},
    data/.style={draw=black!50, fill=black!3, rounded corners, minimum width=36mm, minimum height=7mm, align=center, font=\small\itshape, text=black!70},
    decision/.style={draw, diamond, minimum width=28mm, minimum height=10mm, align=center, font=\small, aspect=2},
    excl/.style={draw, rounded corners, fill=red!10, minimum width=28mm, minimum height=7mm, align=center, font=\small},
    arrow/.style={-{Stealth}, thick}
]
\node[process] (collect) {Data Collection(\S3.1)};
\node[data, below=of collect] (candidates) {Candidate Videos};
\node[process, below=of candidates] (tsp) {TSP Annotation(\S3.2)};
\node[decision, below=of tsp, yshift=-2mm] (decision) {TSP-2/3/4?};
\node[excl, left=of decision, xshift=-18mm] (excluded) {TSP-0 / TSP-1\\ Excluded};
\node[process, below=of decision, yshift=-2mm] (diarization) {Audio-Based Speaker Diarization(\S3.3)};
\node[process, below=of diarization] (annotation) {LLM-Based Annotation(\S3.4)};
\node[data, below=of annotation] (corpus) {Final Corpus};

\draw[arrow] (collect) -- (candidates);
\draw[arrow] (candidates) -- (tsp);
\draw[arrow] (tsp) -- (decision);
\draw[arrow] (decision) -- node[above, font=\footnotesize]{Yes} (diarization);
\draw[arrow] (decision) -- node[above, font=\footnotesize]{No} (excluded);
\draw[arrow] (diarization) -- (annotation);
\draw[arrow] (annotation) -- (corpus);
\end{tikzpicture}
\caption{Four-stage corpus construction pipeline. Data artifacts (grey italic) are distinguished from processing stages (white boxes). Nodes are labeled with the corresponding subsection of \S3.}
\label{fig:tsp_workflow}
\end{figure}
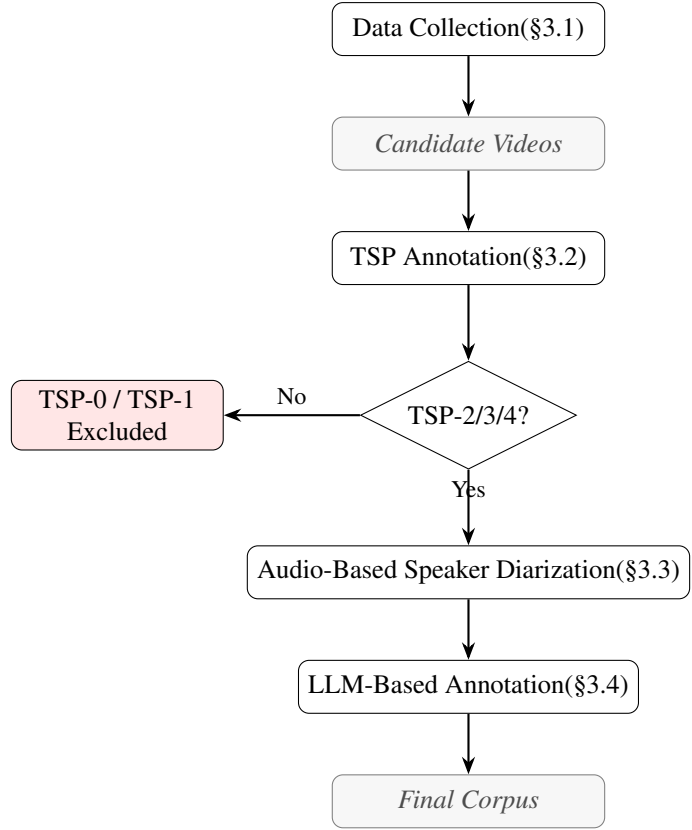

Figure~\ref{fig:tsp_workflow} gives an overview of the corpus construction pipeline. Grey italic nodes denote data artifacts (input videos and the final corpus), while white boxes represent the four processing stages detailed in \S3.1--\S3.4.

\subsection{Data Collection}\label{sec:data_collection}

For each speaker, candidate YouTube videos were retrieved through the yt-dlp\footnote{yt-dlp: \url{https://github.com/yt-dlp/yt-dlp}} search interface using the speaker's name as the primary query, supplemented by the suffixes \texttt{interview}, \texttt{talk}, and \texttt{podcast} to prioritise conversational content. Up to 100 results were collected per query, and results were deduplicated and sorted by upload date. This yielded 2,479 unique candidates across 124 speakers.

Speaker selection followed a stratified design across seven professional domains to ensure representation of distinct discourse communities:

\begin{table}[H]
\centering
\caption{Professional domain coverage in the PDC.}
\label{tab:domains}
\begin{tabular}{lp{0.68\textwidth}}
\toprule
\textbf{Domain} & \textbf{Examples} \\
\midrule
Academia/Economics & Rogoff, Acemoglu, Krugman, Stiglitz, Reinhart, Shiller, Tooze \\
Central Banking/Policy & Powell, Yellen, Bernanke, Greenspan, Lagarde, Draghi, Kashkari \\
Finance/Investing & Dalio, Wood, Ackman, Dimon, Druckenmiller, El-Erian, Marks, Gross \\
Geopolitics/Strategy & Zeihan, Mearsheimer, Bremmer, Kissinger, Rice, McFaul \\
Media/Commentary & Shapiro, Carlson, Peterson, Klein, Harris, Brooks \\
Politics/Government & Trump, Biden, Clinton, Sanders, Macron, Modi, Zelenskyy, Vance \\
Technology/Business & Musk, Altman, Andreessen, Weinstein, Ramaswamy \\
\bottomrule
\end{tabular}
\end{table}

After search, basic availability filtering was applied: automatic English subtitles (VTT) were downloaded via yt-dlp; videos that were unavailable, private, or lacked English captions were removed, and non-English content was excluded based on metadata language tags. Eligible candidates were then sampled evenly across speakers (up to a per-speaker cap, with a fixed random seed for reproducibility) and their audio was downloaded in parallel as m4a files. The key speaker-attribution verification---distinguishing genuine first-person speech from third-party content---is performed manually via the TSP protocol described in \S3.2.

Each selected video retained metadata including YouTube ID, title, upload date, duration, channel name, and language. The temporal span covers 2016--2026 with denser sampling in 2024--2026. We note that the limitations of this selection strategy are discussed in Section~\ref{sec:limitations}.

\subsection{Target Speaker Participation Annotation}\label{sec:tsp}

A critical but often under-documented challenge in building target-speaker corpora from automatically retrieved video sources is that search results frequently yield content \textit{about} the target individual rather than speech \textit{by} them. Existing public datasets rarely provide systematic documentation of how target-speaker presence was verified, creating ambiguity about whether annotated utterances can be reliably attributed. To address this gap, we introduce a manual \textbf{Target Speaker Participation (TSP)} annotation stage prior to downstream processing.

The TSP annotation does not categorize videos by topical content. Rather, it characterizes the participation pattern of the target speaker within each candidate video, serving two purposes: (1) filtering automatically collected videos that are unsuitable for downstream linguistic analysis, and (2) recording the conversational context in which the target speaker appears as metadata for subsequent speaker attribution and corpus stratification.

\subsubsection{TSP Annotation Taxonomy}

We define five mutually exclusive participation categories according to the role of the target speaker in the collected video.

\begin{table}[H]
\centering
\caption{Target Speaker Participation (TSP) annotation taxonomy. Only TSP-2, TSP-3, and TSP-4 are retained for the final corpus.}
\label{tab:tsp_taxonomy}
\begin{tabular}{llp{0.55\textwidth}c}
\toprule
\textbf{Label} & \textbf{Category} & \textbf{Description} & \textbf{Corpus} \\\\
\midrule
TSP-0 & Target Absent & The target speaker produces no identifiable spoken utterance in the video. May appear visually or be mentioned by others. & Excluded \\\\
TSP-1 & Referenced Subject & The target speaker is primarily the subject of discussion or quotation. Brief speech clips may appear but are insufficient for analysis. & Excluded \\\\
TSP-2 & Multi-party Discussion & The target speaker is one of multiple guest participants in a conversation with $\geq 3$ sustained speakers (panels, debates, multi-guest podcasts). The target shares speaking time with other guests. & Included \\\\\\\\
TSP-3 & Sole-Respondent Interview & The target speaker is the primary interviewee, questioned by one or more hosts. The target is the sole guest respondent and contributes substantial speech. & Included \\\\\\\\
TSP-4 & Solo Presentation & The target speaker is the only sustained speaker (speeches, lectures, keynotes, monologues). & Included \\\\
\bottomrule
\end{tabular}
\end{table}

Only videos assigned to \textbf{TSP-2}, \textbf{TSP-3}, and \textbf{TSP-4} are retained for subsequent speaker attribution and valence-modality annotation. TSP-0 and TSP-1 videos are excluded from the corpus.

\subsubsection{Annotation Guidelines}

To ensure annotation consistency, we define several operational rules:

\begin{itemize}[nosep]
  \item \textbf{Target speech requirement.} The target speaker must produce sufficient continuous speech to support downstream linguistic analysis. Videos containing only short quotations, isolated clips, or incidental remarks are classified as TSP-1 rather than as valid target-speaker data.
  \item \textbf{Primary conversational role.} The assigned category is determined by the dominant interaction pattern rather than short introductory or concluding segments. For example, a one-hour interview preceded by a one-minute news introduction is annotated as TSP-3.
  \item \textbf{Target speaker's conversational role.} Videos where the target speaker is the only speaker are labeled TSP-4. Videos where the target speaker is the sole interviewee---interviewed by one or more hosts---are labeled TSP-3. The number of hosts is irrelevant; the key criterion is whether the target is the only respondent. Videos where the target speaker is one of multiple guest participants sharing the floor are labeled TSP-2.
  \item \textbf{Referenced versus participating.} If the target individual is primarily discussed by other speakers---regardless of whether photographs, archival footage, or short quoted speech are presented---the video is annotated as TSP-1 unless the target speaker actively participates in the conversation.
\end{itemize}

\subsubsection{TSP Design Rationale}

A natural question is whether five categories are necessary: the core filtering task requires only a binary decision (include vs.\ exclude), so why not collapse the taxonomy to two levels? We retain five categories because they serve two distinct purposes beyond filtering.

First, distinguishing TSP-0 (complete absence) from TSP-1 (referenced subject) records \textit{why} a video was excluded, enabling diagnosis of retrieval quality across speakers and domains. TSP-0 captures retrieval noise while TSP-1 captures a content mismatch: the video is \textit{about} the speaker but contains no analyzable speech \textit{by} them.

Second, TSP-2, TSP-3, and TSP-4 are not filtering labels but \textit{conversational metadata}. They record whether the target speaker appears as one of multiple guests in a multi-party discussion, as the sole respondent in an interview, or as the only speaker in a solo presentation. This structural information supports downstream analyses: researchers may wish to restrict their study to sole-respondent interviews only, or to compare modality usage between solo speeches and conversational settings. Collapsing these into a single ``included'' label would discard this information.

The taxonomy follows two design principles: it is \textbf{mutually exclusive} (every candidate video can be assigned exactly one label based on the target speaker's dominant participation pattern) and \textbf{exhaustive} with respect to common interview-format content on YouTube (the five categories cover the full spectrum from complete absence to total dominance). The choice of five categories reflects a deliberate trade-off: too few collapse meaningfully distinct patterns, while too many create boundary ambiguity and reduce inter-annotator agreement. The most consequential boundary---TSP-1 (Referenced Subject) vs.\ included categories---is operationalized as continuous participation sufficient to produce at least several paragraphs of attributable text, rather than isolated soundbites or archival clips.

\subsubsection{Inter-Annotator Agreement}

To validate the reliability of the TSP taxonomy, two annotators independently labeled a random sample of 188 candidate videos. Inter-annotator agreement, measured using Cohen's $\kappa$~\cite{artstein_iaa}, was substantial\footnote{Following the interpretation bands of Landis \& Koch (1977): $>$0.80 = almost perfect, 0.61--0.80 = substantial, 0.41--0.60 = moderate, 0.21--0.40 = fair, $\leq$0.20 = slight.} ($\kappa = 0.616$, observed agreement 70.2\%) for the five-way classification and $\kappa = 0.690$ (observed agreement 85.1\%) for the binary inclusion/exclusion decision (TSP-0/1 vs.\ TSP-2/3/4). The largest disagreement source was detection of target-speaker presence (TSP-0 vs.\ included categories, 20 of 56 disagreements), rather than the TSP-1/TSP-2 speech-threshold boundary (only 3 disagreements). All disagreements were resolved through discussion. Detailed results appear in Section~\ref{sec:evaluation}.

This verification stage substantially improves corpus quality by ensuring that all retained videos contain sufficient analyzable speech from the target speaker, while preserving metadata describing the conversational structure of each recording. The TSP taxonomy transforms an implicit curation step---``we manually removed irrelevant videos''---into a replicable annotation protocol with documented categories and decision rules.

\subsection{Audio-Based Speaker Diarization}\label{sec:diarization}

A central methodological challenge in building a corpus from YouTube interviews is speaker diarization: without it, interviewer questions, host narration, and target-speaker responses become indistinguishable, contaminating downstream analyses. We experimented with two approaches---text-only (LLM-based classification) and audio-based---and found that acoustic speaker-turn evidence is essential.

\paragraph{Text-only versus audio diarization.}
The text-only approach segments VTT transcripts into sentence windows and submits each to an LLM (DeepSeek-V4-Flash) for binary guest/interviewer classification. The audio approach uses a commercial ASR service (AssemblyAI Best tier) with native speaker diarization, post-processed to identify the target speaker by word-count dominance with LLM fallback for ambiguous cases. On 35 Ray Dalio videos, the text method produced guest ratios tightly clustered near 100\% (variance 0.002), systematically failing to detect interviewer speech. The audio method showed genuine variance (range 37--173\%, variance 0.04), detecting multiple speakers in 24/35 files. Ground truth validation (3 files, 60 manually annotated utterances) confirmed 91.7\% utterance-level accuracy ($\kappa = 0.82$) for the audio approach---comparable to published diarization results on conversational speech where diarization error rates of 15--30\% are typical~\cite{pyannote}. Text-only diarization fails because without acoustic turn signals the LLM defaults to the target speaker for any ambiguous utterance.

\paragraph{Local diarization pipeline.}
Based on these findings, we implemented a fully local pipeline combining \textbf{faster-whisper}\footnote{faster-whisper: \url{https://github.com/SYSTRAN/faster-whisper}} (large-v3) for ASR---a reimplementation of OpenAI's Whisper~\cite{whisper} with CTranslate2 inference---with \textbf{pyannote/speaker-diarization-3.1}~\cite{pyannote} for acoustic speaker separation on a single consumer GPU. The pipeline: (1) downloads and converts audio to 16\,kHz mono WAV; (2) transcribes with faster-whisper; (3) produces speaker-turn annotations with pyannote (with automatic audio chunking to avoid GPU memory exhaustion); (4) fuses the outputs by assigning each whisper segment the pyannote speaker with the longest time-overlap. Target-speaker identification uses a rule-based decision tree---single-speaker files assign directly, two-speaker files use a word-count dominance threshold ($r < 0.3$), and ambiguous or multi-speaker cases fall back to an LLM prompted with the target speaker's name. The pipeline is checkpointed and controlled by a simple file list for incremental processing.

\paragraph{Cross-pipeline validation.} To validate that our local pipeline produces reliable transcripts, we compare it against AssemblyAI (Best tier) on 212 overlapping files spanning 52 speakers across all seven domains. Transcript agreement is high: average Jaccard similarity 0.706, median 0.744, with 84.1\% word overlap. Vocabulary quality is comparable (filler ratios 3.16\% vs.\ 3.36\%, TTR 0.345 vs.\ 0.341). Only 20 files (9.4\%) fall below Jaccard 0.50 (a diagnostic threshold below which transcripts were manually inspected), all due to the two pipelines selecting \emph{different} speakers as the target in multi-participant panels rather than transcription failures.

\paragraph{Evidence scope for local pipeline reliability.}
We emphasize an important distinction: the cross-pipeline validation described above measures \textit{transcript consistency} (do two ASR pipelines produce similar word sequences?), not \textit{speaker attribution accuracy} (does each utterance in the final corpus belong to the claimed speaker?). The Jaccard similarity of 0.706 and 84.1\% word overlap confirm that our local Whisper + pyannote pipeline produces transcripts comparable to a commercial alternative, but they do not directly validate that the target-speaker identification step (the rule-based decision tree and LLM fallback) correctly assigns each utterance to the intended speaker. The 60-utterance ground-truth validation on AssemblyAI output (91.7\% accuracy) provides a reference point, but this ground truth was collected on the commercial pipeline, not on the local pipeline released with the corpus. We therefore characterize the local pipeline's speaker attribution as \textit{transcript-consistent but not yet directly validated for speaker attribution accuracy}. A systematic human evaluation of speaker-attribution accuracy across the full corpus---stratified by TSP category, speaker count, and domain---is planned as immediate follow-up work. In the interim, we release the full pipeline code so that users can replicate, inspect, and independently evaluate the speaker attribution step.

\paragraph{Diarization quality assessment.} We evaluate diarized transcripts along four dimensions---text quality (speaking rate, filler-word rate, short-segment proportion), speaker separation (speaker count, UNKNOWN proportion), temporal completeness (audio coverage, gap proportion), and anomaly detection (repeated trigrams, hallucination-like patterns)---each normalized to 0--100 and combined into a weighted composite. Applying this framework to all 1,088 diarized transcripts from the local pipeline, 998 (91.7\%) are rated GOOD ($\ge$80) or FAIR ($\ge$60). A total of 90 files (8.3\%) carry POOR ($\ge$40) or BAD ($<$40) flags and are discarded before downstream analyses.

\subsection{LLM-Based Annotation}\label{sec:llm_annotation}

After punctuation-based sentence splitting, each sentence is submitted to a zero-shot dual-dimension classification task: affective valence (positive/negative/neutral) and epistemic modality (emphatic/hedged/neutral). The two dimensions are orthogonal by design, capturing both \textit{what} a speaker says and \textit{how} assertively they say it. The classifier is instructed to prefer neutral when uncertain and to output both labels in structured JSON. This dual-dimension scheme was validated in prior work~\cite{chen_certainty_artifact}; we adopt it here for reproducibility.

For the released corpus, all eligible sentences are annotated rather than sampled. Transcripts are batched (50 sentences per batch) and submitted to DeepSeek-V4-Flash (\texttt{temperature=0}, thinking disabled) as the primary annotation model. To assess cross-provider label stability, a stratified 1\% sample ($N = 1{,}863$) was independently re-annotated by GPT-5.5 (reasoning effort = \texttt{none}). Agreement between the two providers is reported in Section~\ref{sec:llm_agreement}.


\section{The Public Discourse Corpus}
\subsection{Corpus Statistics}

Table~\ref{tab:corpus_stats} summarizes the overall scope of the PDC and Table~\ref{tab:domain_breakdown} provides the per-domain breakdown of speakers, videos, and sentences.

\begin{table}[H]
\centering
\caption{PDC corpus scope.}
\label{tab:corpus_stats}
\begin{tabular}{@{}lr@{}}
\toprule
\textbf{Statistic} & \textbf{Value} \\
\midrule
Professional domains & 7 \\
Speakers (candidate pool) & 124 \\
Candidate videos (search) & 2,479 \\
Downloaded (audio retrieved) & 1,449 \\
Videos passed TSP and diarized & 1,088 \\
Videos in final corpus & 998 \\
Speakers in final corpus & 100 \\
Sentences & 186,642 \\
Words (in sentences) & 3,113,156 \\
Mean words / sentence & 16.7 \\
\bottomrule
\end{tabular}
\end{table}

\begin{table}[H]
\centering
\caption{Per-domain breakdown of speakers, videos, and sentences in the PDC.}
\label{tab:domain_breakdown}
\begin{tabular}{@{}lrrr@{}}
\toprule
\textbf{Domain} & \textbf{Speakers} & \textbf{Videos} & \textbf{Sentences} \\
\midrule
Academia/Economics & 15 & 124 & 26,151 \\
Central Banking/Policy & 9 & 84 & 10,202 \\
Finance/Investing & 21 & 215 & 41,319 \\
Geopolitics/Strategy & 14 & 147 & 32,385 \\
Media/Commentary & 16 & 216 & 47,291 \\
Politics/Government & 22 & 183 & 25,348 \\
Technology/Business & 3 & 29 & 3,946 \\
\bottomrule
\end{tabular}
\end{table}

Each included video is verified through TSP (\S3.2) to contain genuine first-person speech. Sentences average 16.7 words ($\sigma = 10.4$), with 72.4\% in the 6--20 range. Speaker contributions follow a long-tail distribution (median 1,300, IQR 539--2,865), characteristic of real-world public discourse. Table~\ref{tab:label_distribution} shows the joint distribution of valence and modality across the 186,642 annotated sentences: valence is balanced with a slight negative skew (34.7\% negative, 24.9\% positive), and modality is predominantly neutral (58.0\%).

\begin{table}[H]
\centering
\caption{Marginal and joint distribution of valence and modality labels. Each cell shows count and row percentage.}
\label{tab:label_distribution}
\begin{tabular}{lrrrr}
\toprule
\textbf{Valence $\setminus$ Modality} & \textbf{Emphatic} & \textbf{Neutral} & \textbf{Hedged} & \textbf{Total} \\
\midrule
Positive     & 14,396 (31.0\%) & 21,415 (46.1\%) & 10,667 (22.9\%) & 46,478 (24.9\%) \\
Negative     & 16,562 (25.6\%) & 32,789 (50.6\%) & 15,441 (23.8\%) & 64,792 (34.7\%) \\
Neutral      &  2,965 ( 3.9\%) & 54,016 (71.7\%) & 18,391 (24.4\%) & 75,372 (40.4\%) \\
\midrule
Total        & 33,923 (18.2\%) & 108,220 (58.0\%) & 44,499 (23.8\%) & 186,642 \\
\bottomrule
\end{tabular}
\end{table}

\subsection{Dataset Structure}\label{sec:dataset_structure}

The PDC release consists of three JSONL files, joined by the \texttt{video\_id} and \texttt{speaker} keys:

\paragraph{\texttt{anno.jsonl} (186,642 records).} The primary annotation file. Each record contains six fields:
\begin{itemize}[nosep]
  \item \texttt{video\_id} --- YouTube video identifier, joining to \texttt{videos.jsonl}.
  \item \texttt{speaker} --- speaker slug (e.g., \texttt{ray\_dalio}), joining to \texttt{speakers.jsonl}.
  \item \texttt{sentence\_index} --- zero-based position within the video transcript.
  \item \texttt{sentence} --- cleaned, diarized target-speaker text.
  \item \texttt{valence} --- affective label (\texttt{positive} / \texttt{neutral} / \texttt{negative}).
  \item \texttt{modality} --- epistemic modality label (\texttt{emphatic} / \texttt{neutral} / \texttt{hedged}).
\end{itemize}

\paragraph{\texttt{speakers.jsonl} (100 records).} Speaker metadata, keyed by \texttt{speaker}:
\begin{itemize}[nosep]
  \item \texttt{speaker} --- slug (primary key).
  \item \texttt{name} --- display name.
  \item \texttt{domain} --- professional domain (one of seven categories).
  \item \texttt{n\_videos} --- number of videos contributed.
  \item \texttt{video\_ids} --- list of video IDs for this speaker.
\end{itemize}

\paragraph{\texttt{videos.jsonl} (998 records).} Video metadata, keyed by \texttt{video\_id}:
\begin{itemize}[nosep]
  \item \texttt{video\_id} --- YouTube identifier (primary key).
  \item \texttt{title} --- video title.
  \item \texttt{channel} --- channel name.
  \item \texttt{upload\_date} --- upload date (YYYY-MM-DD).
  \item \texttt{duration} --- video duration (MM:SS or HH:MM:SS).
  \item \texttt{description} --- truncated video description.
\end{itemize}

All downstream aggregates (per-speaker, per-domain, cross-dimension correlations) are derivable from these three files through standard join operations.

\subsection{Illustrative Analysis: Valence--Modality Correlation}\label{sec:illustrative}

To demonstrate one type of research question the PDC enables---rather than as a core contribution of the dataset---we present a brief cross-dimension correlation analysis. The dual-dimension annotation reveals that valence and modality are largely orthogonal at the corpus level. We quantify the coupling between dimensions using the phi coefficient ($\varphi$) computed on binary indicators (e.g., \texttt{valence=positive} vs.\ \texttt{modality=emphatic}) for each sentence.

\begin{table}[H]
\centering
\caption{Cross-dimension phi correlations, overall and by domain (mean per-speaker $\varphi$).}
\label{tab:cross_dim}
\begin{tabular}{@{}lrrrr@{}}
\toprule
\textbf{Domain} & \textbf{$\varphi$(pos,emp)} & \textbf{$\varphi$(neg,emp)} & \textbf{$\varphi$(pos,hed)} & \textbf{$\varphi$(neg,hed)} \\
\midrule
Overall (pooled)       & 0.191 & 0.140 & $-0.012$ & $-0.000$ \\
\midrule
Academia/Economics     & 0.153 & 0.162 & $-0.002$& $-0.023$\\
Central Banking/Policy & 0.225 & 0.110 & 0.009   & 0.028   \\
Finance/Investing      & 0.230 & 0.101 & 0.009   & 0.014   \\
Geopolitics/Strategy   & 0.113 & 0.168 & 0.030   & $-0.007$\\
Media/Commentary       & 0.171 & 0.163 & $-0.019$& $-0.024$\\
Politics/Government    & 0.211 & 0.125 & $-0.020$& $-0.033$\\
Technology/Business    & 0.519 & 0.056 & $-0.054$& $-0.047$\\
\bottomrule
\end{tabular}
\end{table}

The descriptive pattern is an asymmetry between positive and negative valence in their coupling with emphatic modality. Across most domains, positive statements are more likely to be emphatic than negative statements (pooled $\varphi(\text{pos},\text{emp}) = 0.191$ vs.\ $\varphi(\text{neg},\text{emp}) = 0.140$). Technology/Business is an outlier ($\varphi = 0.519$), but this domain contains only 3 speakers and 29 videos (see Table~\ref{tab:domain_breakdown}), so the estimate is driven by a small sample and should be interpreted with caution. Geopolitics and Academia show the reverse pattern (negative statements more often emphatic than positive ones), though the effect sizes are modest ($\varphi < 0.17$). Correlations involving hedged modality are near zero across all domains, indicating that hedging is largely orthogonal to valence in this corpus. We do not attribute these observations to discourse styles or rhetorical norms; they are presented as descriptive patterns that may motivate confirmatory studies with controlled designs.

We emphasize that this analysis is illustrative: it shows one way the PDC can be used, but the full space of possible analyses---including speaker-level regression models, temporal trend analysis, and cross-domain comparative studies---is substantially larger. The corpus is designed to support such analyses, not to preempt them.

\section{Dataset Release and Documentation}

The PDC is released under a research-use license. The release comprises four components:

\paragraph{Annotated corpus.} The dataset comprises the three files described in \S4.2: \texttt{anno.jsonl} (186,642 annotated sentences), \texttt{speakers.jsonl} (100 speakers with domain labels), and \texttt{videos.jsonl} (998 videos with title, channel, upload date, and duration metadata).

\paragraph{Validation data.} A stratified 1\% sample ($N = 1{,}863$) was independently re-annotated by GPT-5.5, yielding the cross-provider agreement statistics reported in \S6.3. These annotations and the sample manifest are included.

\paragraph{Annotation tooling.} A lightweight web-based annotation application supports three manual annotation workflows used during corpus construction. \textbf{TSP video annotation} presents each candidate video alongside its YouTube metadata and accepts a five-level TSP rating (0--4), with a ``skip'' option for ambiguous cases; grades are persisted per annotator to enable inter-annotator agreement computation (\S6.1). \textbf{Diarization annotation} samples sentences from diarized transcripts and supports per-speaker filtering, allowing reviewers to spot-check for diarization errors (e.g., misattributed turns, residual host speech). \textbf{LLM annotation review} displays sentences with their LLM-assigned valence and modality labels and allows human reviewers to confirm or override each label in a side-by-side interface, directly supporting the human gold-standard evaluation described in \S6.3. All three tools run locally with no external service dependencies.

\paragraph{Reproducible pipeline.} All processing code is released as open-source Python scripts, organized into modules mirroring the pipeline stages: YouTube candidate search and audio download (01\_collect), local diarization via faster-whisper + pyannote with quality assessment (02\_asr), sentence segmentation and filtering (03\_sentence\_splitting), and LLM-based annotation with checkpointed batch processing (04\_annotation). A central configuration module (\texttt{config.py}) defines the speaker--domain mapping and all pipeline parameters. The local diarization stage runs on a consumer GPU without proprietary API keys.

Documentation follows the \textit{Datasheets for Datasets} framework \cite{gebru_2021}.

\paragraph{Availability.} The annotated corpus is available at \url{https://huggingface.co/datasets/ictchenbo/public-discourse-corpus}. The annotation tools and processing pipeline are released as open-source software at \url{https://github.com/ictchenbo/pdc-construction-pipeline}.

\section{Evaluation}\label{sec:evaluation}

We evaluate the PDC construction pipeline along three dimensions: TSP annotation reliability, speaker diarization accuracy, and annotation quality. Primary LLM annotation has been completed for all 186,642 sentences across 100 speakers using DeepSeek-V4-Flash. Cross-provider validation against GPT-5.5 on a stratified 1\% sample is reported in Section~\ref{sec:llm_agreement}.

\subsection{TSP Inter-Annotator Agreement}

Two annotators independently labeled a random sample of 188 candidate videos using the TSP taxonomy. Agreement was substantial (Cohen's $\kappa = 0.616$, observed agreement 70.2\%) for the five-way classification and $\kappa = 0.690$ (observed agreement 85.1\%) for the binary inclusion decision (TSP-0/1 vs.\ TSP-2/3/4).

The confusion matrix (Table~\ref{tab:tsp_confusion}) reveals two main disagreement sources. First, presence detection: 20 of 56 disagreements arise when one annotator assigns TSP-0 (target absent) while the other assigns an included category (TSP-2/3/4). These typically involve videos where the target speaker appears only briefly, and annotators differ on whether the speech is sustained enough for downstream analysis. Second, interview-format boundaries: 10 cases fall at the TSP-2/TSP-3 boundary (multi-party discussion vs.\ sole-respondent interview) and 9 at the TSP-3/TSP-4 boundary (interview vs.\ solo presentation). These reflect genuine ambiguity in conversational format---for example, whether a podcast with two hosts interviewing a single guest is best characterized as ``multi-party'' or ``sole-respondent interview.''

The most consequential boundary for corpus construction---TSP-1 (referenced subject) vs.\ included categories---accounts for only 3 of 56 disagreements, indicating that annotators reliably distinguish between insufficient and sufficient target-speaker speech. Per-category agreement is lowest for TSP-0 (42.9\%) and highest for TSP-1 (88.6\%), confirming that the primary challenge is detecting target-speaker presence rather than discriminating among participation patterns. All disagreements were resolved through discussion and the resulting consensus labels are included in the released corpus.

\begin{table}[H]
\centering
\caption{Confusion matrix for TSP inter-annotator agreement (Annotator A rows, Annotator B columns). Diagonal shows agreement counts.}
\label{tab:tsp_confusion}
\begin{tabular}{@{}lrrrrr@{}}
\toprule
\textbf{A $\backslash$ B} & \textbf{TSP-0} & \textbf{TSP-1} & \textbf{TSP-2} & \textbf{TSP-3} & \textbf{TSP-4} \\
\midrule
TSP-0 & 15 & 4 & 4 & 11 & 1 \\
TSP-1 & 3  & 39 & 0 & 1  & 1 \\
TSP-2 & 2  & 0  & 7 & 3  & 0 \\
TSP-3 & 1  & 2  & 7 & 37 & 6 \\
TSP-4 & 1  & 4  & 2 & 3  & 34 \\
\bottomrule
\end{tabular}
\end{table}

\subsection{Speaker Diarization Accuracy}

We evaluate two diarization pipelines: (1) the cloud-based AssemblyAI Best tier, and (2) our local Whisper + pyannote pipeline. On the 35-file Ray Dalio comparison set, AssemblyAI achieved 91.7\% utterance-level accuracy ($\kappa = 0.82$) against human ground truth (60 manually annotated utterances across 3 files). As discussed in Section~\ref{sec:diarization}, this ground truth applies to the AssemblyAI pipeline, not to the local pipeline, and the sample is small (3 files, 60 utterances). The local pipeline's transcript consistency with AssemblyAI is strong (Jaccard 0.706, 84.1\% word overlap on 212 files), but we do not yet have direct human validation of the local pipeline's speaker-attribution accuracy. The multi-dimensional quality assessment framework described in Section~\ref{sec:diarization} rates 91.7\% of diarized transcripts (998/1,088) as GOOD or FAIR; the remaining 90 files (8.3\%) are discarded before downstream analyses. We release the full pipeline code and encourage independent evaluation of speaker-attribution accuracy on domain-specific subsets.

\subsection{LLM Annotation Quality}\label{sec:llm_agreement}

To assess the reliability of LLM-based annotation, we compare the primary DeepSeek-V4-Flash annotations against GPT-5.5 on a stratified 1\% sample of 1,863 sentences, drawn proportionally across all seven domains (per-domain $N$: Media 472, Finance 413, Geopolitics 323, Academia 261, Politics 253, Central Banking 102, Technology 39). Both models are prompted with the same zero-shot dual-dimension classification task described in Section~\ref{sec:llm_annotation}; DeepSeek-V4-Flash uses thinking = \texttt{disabled} and GPT-5.5 uses reasoning effort = \texttt{none}.

\begin{table}[H]
\centering
\caption{Cross-provider agreement: DeepSeek-V4-Flash vs.\ GPT-5.5 on a stratified 1\% sample ($N = 1{,}863$).}
\label{tab:llm_agreement}
\begin{tabular}{@{}lrr@{}}
\toprule
\textbf{Metric} & \textbf{Valence} & \textbf{Modality} \\
\midrule
Exact agreement & 75.9\% & 72.5\% \\
Cohen's $\kappa$ (unweighted) & 0.63 & 0.53 \\
Polarity flips (neg $\leftrightarrow$ pos) & 30 & --- \\
\bottomrule
\end{tabular}
\end{table}

Cross-provider agreement falls in the moderate-to-substantial range: valence agreement is 75.9\% ($\kappa = 0.63$), modality agreement is 72.5\% ($\kappa = 0.53$). Table~\ref{tab:domain_kappa} reports per-domain agreement. Excluding Technology/Business ($N = 39$, too small for reliable per-domain estimates), Central Banking shows the highest cross-provider consistency for both dimensions, while Academia shows the lowest modality agreement ($\kappa = 0.46$). The overall pattern is consistent with published findings on LLM annotation variability~\cite{gilardi_2023, ziems_2024}. The lower agreement for modality reflects the inherent difficulty of epistemic modality classification: distinguishing hedged from neutral statements requires pragmatic inference that is more sensitive to provider-specific decision boundaries than polarity detection. These results underscore that single-provider LLM annotations represent one measurement perspective; the released multi-provider validation sample enables users to bound label uncertainty in their own analyses.

\begin{table}[H]
\centering
\caption{Per-domain cross-provider agreement (DeepSeek-V4-Flash vs.\ GPT-5.5). $\kappa$ = unweighted Cohen's Kappa.}
\label{tab:domain_kappa}
\begin{tabular}{@{}lrrrrr@{}}
\toprule
\textbf{Domain} & \textbf{N} & \textbf{V Agr.} & \textbf{V $\kappa$} & \textbf{M Agr.} & \textbf{M $\kappa$} \\
\midrule
Media/Commentary     & 472 & 77.8\% & 0.66 & 71.4\% & 0.52 \\
Finance/Investing    & 413 & 75.5\% & 0.62 & 71.9\% & 0.50 \\
Geopolitics/Strategy & 323 & 74.0\% & 0.59 & 74.0\% & 0.57 \\
Academia/Economics   & 261 & 75.5\% & 0.60 & 70.1\% & 0.46 \\
Politics/Government  & 253 & 73.9\% & 0.60 & 73.1\% & 0.55 \\
Central Banking/Policy & 102 & 79.4\% & 0.66 & 77.5\% & 0.60 \\
Technology/Business  & 39  & 79.5\% & 0.67 & 76.9\% & 0.64 \\
\midrule
Overall              & 1,863 & 75.9\% & 0.63 & 72.5\% & 0.53 \\
\bottomrule
\end{tabular}
\end{table}

\paragraph{Human gold standard (ongoing work).} To establish an absolute upper bound on annotation reliability, a random sample of approximately 400 sentences (stratified by disagreement type) is being manually annotated by two trained human judges. LLM-vs.-human agreement will ground the cross-provider comparison reported above and enable accuracy-based evaluation rather than agreement-only analysis. This work is currently in progress; full results will be incorporated in a future revision.

\section{Limitations}

\label{sec:limitations}
\begin{enumerate}
  \item \textbf{Speaker attribution validation:} The local Whisper + pyannote pipeline has been validated for transcript consistency against AssemblyAI (Jaccard 0.706, 84.1\% word overlap on 212 files), but direct human evaluation of speaker-attribution accuracy has not yet been conducted. The existing ground truth (60 utterances across 3 Ray Dalio files, 91.7\% accuracy) applies to the AssemblyAI pipeline, not to the local pipeline. A systematic human evaluation of local-pipeline speaker attribution, stratified by TSP category, speaker count, and domain, is planned as follow-up work.
  \item \textbf{LLM annotation reliability:} All released annotations are produced by a single model (DeepSeek-V4-Flash). Cross-provider validation against GPT-5.5 on a 1\% sample ($N = 1{,}863$) shows moderate agreement ($\kappa = 0.63$ for valence, 0.53 for modality), but spans only two providers and one sample. Human gold-standard annotation is in progress and will provide an absolute accuracy ceiling in a future revision.
  \item \textbf{Corpus coverage:} The corpus is limited to English-language public-figure interviews sourced from YouTube. Video selection involves researcher judgment (format clarity, audio quality, third-person screening) that may not be perfectly replicable. Findings may not generalize to languages with different modality-marking systems.
  \item \textbf{Computational requirements:} The local diarization pipeline requires a consumer GPU, though it eliminates per-file API costs entirely. Researchers without GPU access may substitute a cloud diarization service. Full pipeline code is released to facilitate both approaches.
\end{enumerate}

\section{Conclusion}

We introduce the \textbf{Public Discourse Corpus (PDC)}, the first dataset of public-figure interview speech jointly annotated for affective valence and epistemic modality. The corpus spans 998 videos from 100 speakers across seven professional domains, comprising 186,642 sentences (3.1 million words). The PDC is the primary contribution: it fills a gap in available resources by providing paired valence--modality annotations on speaker-attributed utterances at a scale and domain diversity that enables previously impossible research on the interaction between evaluative language and epistemic commitment.

To ensure that all retained videos contain analyzable speech from the intended speaker, we develop \textbf{Target Speaker Participation (TSP)}, a five-category annotation taxonomy that systematically verifies target-speaker presence in automatically retrieved videos. TSP is a key methodological contribution: it addresses the general ``content about vs.\ speech by'' problem that affects any corpus built from media sources, is validated through documented inter-annotator agreement ($\kappa = 0.616$), and is released as a reusable annotation protocol. Target-speaker turns are separated from interviewer and third-party speech through an \textbf{audio-first diarization pipeline} combining local Whisper ASR with pyannote speaker separation; the pipeline offers stronger evidence for speaker attribution than text-only heuristics and is released as open-source software.

The annotated corpus is available at \url{https://huggingface.co/datasets/ictchenbo/public-discourse-corpus}, the complete processing pipeline at \url{https://github.com/ictchenbo/pdc-construction-pipeline}, and the annotation tool at \url{https://github.com/ictchenbo/pdc-annotator}. We expect these resources to support research on the relationship between evaluative content and epistemic commitment in public discourse, while the TSP methodology and audio-first pipeline offer reusable tools for any project that constructs speaker-attributed corpora from automatically retrieved media sources.

\section*{Ethics Statement}
The PDC consists of publicly available YouTube interviews featuring public figures. All content was already in the public domain at the time of collection. We do not redistribute video or audio; only cleaned, diarized text transcripts and their annotations are released. Speaker names are preserved because the corpus's value lies in its speaker-attributed structure, but we encourage users to consider ethical implications of individual-level analysis, particularly for non-public-figure applications of the methodology.


\end{document}